\documentclass{article}

    \usepackage[nonatbib , preprint]{neurips_2019}

\usepackage[utf8]{inputenc} 
\usepackage[T1]{fontenc}    
\usepackage{hyperref}       
\usepackage{url}            
\usepackage{booktabs}       
\usepackage{amsfonts}       
\usepackage{nicefrac}       
\usepackage{microtype}      
\usepackage{graphicx}
\usepackage{amssymb}
\usepackage{amsmath}

\newtheorem{theorem}{Theorem}
\newtheorem{corollary}{Corollary}
\newtheorem{lemma}{Lemma}
\newtheorem{definition}{Definition}

\title{On Approximation Capabilities of ReLU Activation and Softmax Output Layer in Neural Networks}

\author{%
  Behnam Asadi \\
  Department of Electrical Engineering and Computer Science\\
  York University,  4700 Keele Street, Toronto, ON M3J1P3, Canada \\
  \texttt{benasadi@cse.yorku.ca} \\
   \And
  Hui Jiang \\
  Department of Electrical Engineering and Computer Science\\
  York University,  4700 Keele Street, Toronto, ON M3J1P3, Canada \\
  \texttt{hj@cse.yorku.ca} \\
}

\begin{document}

\maketitle

\begin{abstract}
In this paper, we have extended the well-established universal approximator theory to neural networks that use the unbounded ReLU activation function and a nonlinear softmax output layer. We have proved that a sufficiently large neural network using the ReLU activation function can approximate any function in $L^1$ up to any arbitrary precision. Moreover, our theoretical results have shown that a large enough neural network using a nonlinear softmax output layer can also approximate any indicator function in $L^1$, which is equivalent to mutually-exclusive class labels in any realistic multiple-class pattern classification problems. To the best of our knowledge, this work is the first theoretical justification for using the softmax output layers in neural networks for pattern classification. 
\end{abstract}

\section{Introduction}
\label{Intro}

In recent years, neural networks have revived in the field of machine learning as a dominant method for supervised learning. In practice, large-scale neural networks are trained to classify their inputs based on a huge amount of training samples consisting of input-output pairs. Many different types of network structures can be used to specify the architecture for a neural network, such as fully-connected neural networks, recurrent neural networks, and convolutional neural networks.  Fundamentally speaking, all neural networks just attempt to approximate an unknown target function from its input space to output space. Recent experimental results in many application domains have empirically demonstrated the superb power of various types of neural networks in terms of approximating an unknown function. On the other hand, the \textit{universal approximator} theory has been well established for neural networks since several decades ago \cite{Cybernko89,Kurt1991251}. However, these theoretical works were proved based on some early neural network configurations that were popular at that time, such as sigmoidal activation functions and fully-connected linear output layers. Since then, many new  configurations have been adopted for neural networks. For example, a new type of nonlinear activation function, called 
\textit{rectified linear unit} (ReLU) activation, was originally proposed in  \cite{jarrett2009best,nair2010rectified,pmlr-v15-glorot11a}, and then very quickly it has been widely adopted in neural networks due to its superior performance in training deep neural networks. When the ReLU activation was initially proposed, it was quite a surprise to many people due to the fact that the ReLU function is actually unbounded. In addition, as we know, it is a common practice to use a nonlinear \textit{softmax} output layer when neural networks are used for a pattern classification task \cite{bridle-softmax-1990,10.1007/978-3-642-76153-9_28}. The soft-max output layer  significantly changes the range of the outputs of a neural network due to its non-linearity. 
Obviously, the theoretical results in \cite{Cybernko89,Kurt1991251} are not directly applicable to the unbounded ReLU activation function and the nonlinear \textit{softmax} output layers in neural networks. 

In this paper, we study the approximation power of neural networks that use the ReLU activation function and a \textit{softmax} output layer, and have extended the \textit{universal approximator} theory in \cite{Cybernko89,Kurt1991251} to address these two cases. Similar to \cite{Cybernko89,Kurt1991251}, our theoretical results are established based on a simple feed-forward neural network using a single fully-connected hidden layer. Of course, our results can be further extended to more complicated structures since many complicated structures can be viewed as special cases of fully-connected layers. The major results from this work include: i) A sufficiently large neural network using the ReLU activation function can approximate any $L^1$ function up to any arbitrary precision; ii) A sufficiently large neural network using the \textit{softmax} output layer can approximate well any \textit{indicator function} in $L^1$, which is equivalent to mutually exclusive class labels in a realistic multiple-class pattern classification problem.

\section{Related Works}
\label{works}

Under the notion of universal approximators, the approximation capabilities of some types of neural networks have been studied by other authors. For example, 
\textit{K. Hornik} in \cite{Kurt1991251} has shown the power of fully-connected neural networks using a bounded and non-constant activation layer to approximate any function in ${L}^{p}(I_d)$, where $1 \leq p < \infty$ and $I_d=[0,1]^{d}$ stands for the unit hypercube in an $d$-dimensional space. And ${L}^{p}(I_d)$ denotes the space of all functions defined on $I_d=[0,1]^{d}$ such that $\int_{I_d} |f(\mathbf{x})|^{p} d\mathbf{x} < \infty$. He has proved that if the activation function is a bounded and non-constant function, a large enough single hidden layer feed-forward neural network can approximate any function in ${L}^{p}(I_d)$. He has also proved that same network using a  continuous, bounded and non-constant activation function can approximate any function in ${C}(I_d)$, where ${C}(I_d)$ denotes the space of all continuous functions defined in the unit hypercube $I_d$.
Moreover, \textit{G. Cybenco} in \cite{Cybernko89} has proved that any target functions in ${C}(I_d)$ can be approximated by a single hidden layer feed-forward neural network using any continuous sigmoidal activation function. He has also demonstrated (in Theorem 4) that same network with bounded measurable sigmoidal activation function can approximate any function in ${L}^1(I_d)$ where the sigmoidal activation function is defined as a monotonically-increasing function whose limit is 1 when approaching $+\infty$ and is 0  at $-\infty$. In this paper, we will extend this theorem to neural networks using the unbounded ReLU activation function.

There are also other works related to the approximation power of neural networks using the ReLU activation function but we do not use them in our proofs. For example, in \cite{Leshno93}, it has proven that any activation function will ensure a neural network to be a universal approximator if and only if this function is not a polynomial almost everywhere. The results of this paper could be applied to the ReLU activation function but their target function is in ${C}(I_d)$ space. 
In \cite{arora2018understanding}, the authors have shown that every function in ${L}^p(\mathbb{R}_n)$ $(1\leq p < \infty)$ can be approximated in the $L^p$ norm by a ReLU deep neural network using at most $\lceil \log_2 {(n + 1)} \rceil$ hidden layers. Moreover, for $n=1$, any such $L^p$ function can be arbitrarily well approximated by a 2-layer DNN, with some tight approximation bounds on the size of neural networks.

On the other hand, we have not found much theoretical work on the approximation capability of a \textit{softmax} neural network. A \textit{softmax} output layer is simply taken for granted in pattern classification problems. 

\section{Notations}
\label{Notations}

\begin{figure}[h!]
\centering
  \includegraphics[width=0.45\linewidth]{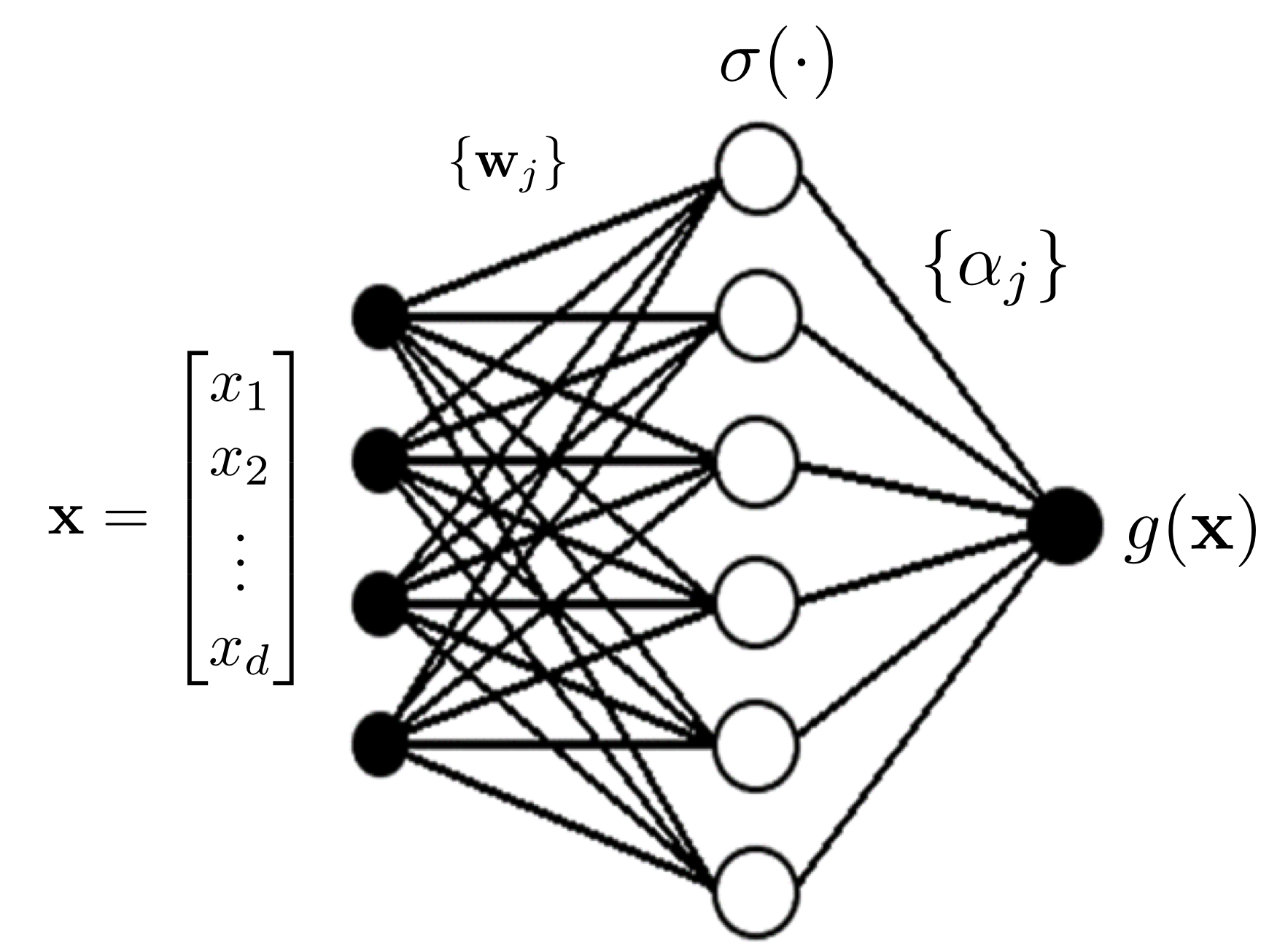}
  \caption{An illustration of a single-output feedforward neural network without any softmax layer.}
  \label{single_output_network}
\end{figure}

We represent the input of a neural network as $ \mathbf{x}= \big[ x_1 \; x_2 \cdots \; x_d \big] \in I_d$, where $I_d$ denotes the $d$-dimensional unit hypercube, i.e. $I_d = [0,1]^d$.
As in Figure \ref{single_output_network},
the output of a single hidden layer feed-forward neural network using a linear output layer
can be represented as a superposition of the activation functions $\sigma(\cdot)$:
$$
g(\mathbf{x})=\sum_{j=1}^{n} \alpha_j \cdot \sigma(\mathbf{w}_j^\intercal \mathbf{x}+b_j)
$$ 
where $g(\mathbf{x}) \in\mathbb{R}$, and 
$\mathbf{w}_j$ denotes the input weights of the $j$-th hidden unit, and $\alpha_j$ stands for the output weight of the $j$-th hidden unit, and $b_j$ is the bias of the $j$-th hidden unit.

\begin{figure}[h!]
\centering
  \includegraphics[width=0.6\linewidth]{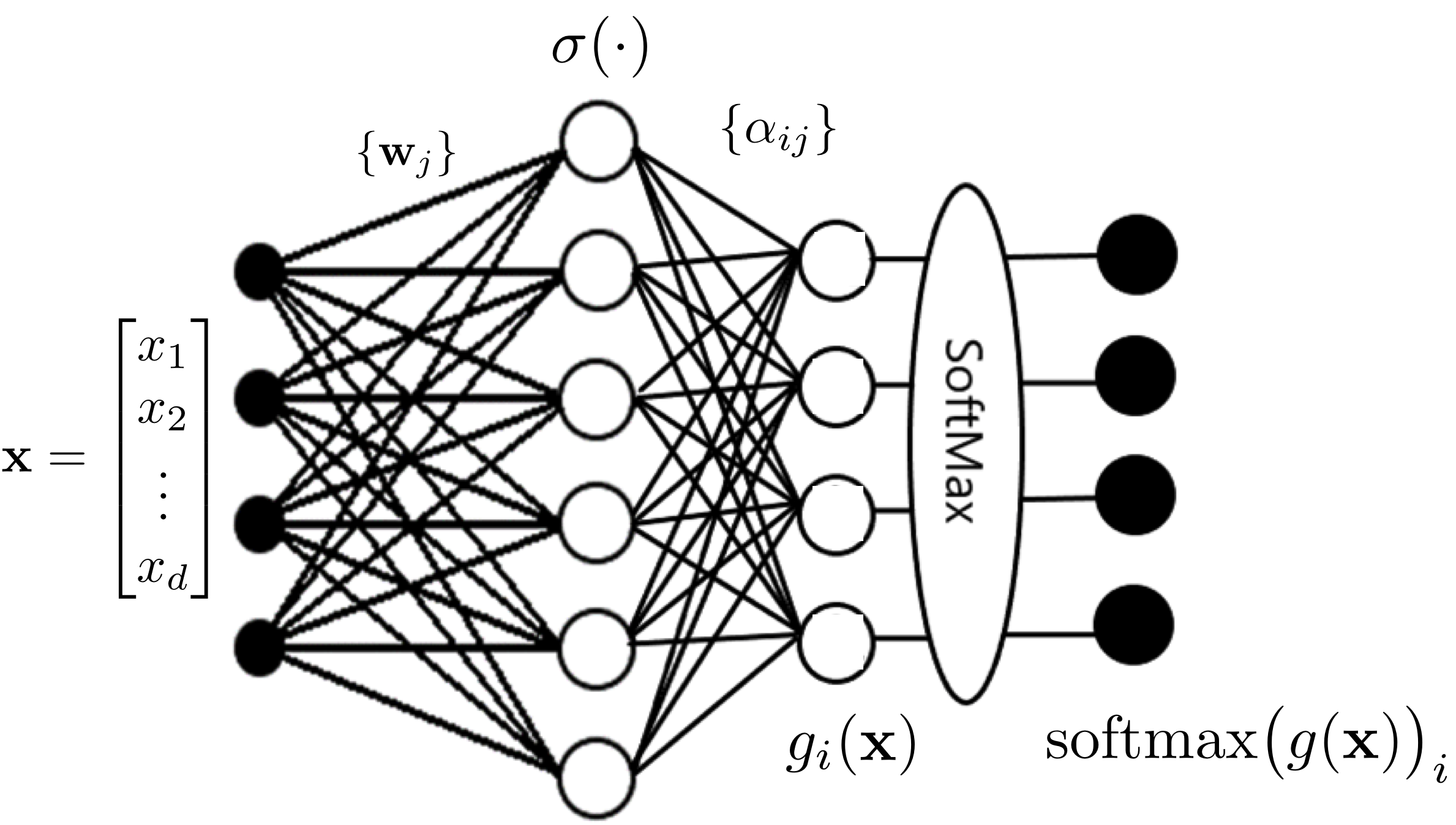}
  \caption{An illustration of a multi-output feedforward neural network with a softmax layer.}
  \label{multi_output_network}
\end{figure}

If a neural network is used for a multiple-class pattern classification problem,  as shown in Figure \ref{multi_output_network}, 
we use the notation $g_i(\mathbf{x})$ for $1 \leq i \leq m$ to represent $i$-th output of the neural network prior to the \textit{softmax} layer. Similarly, we have
$g_i(\mathbf{x})=\sum_{j=1}^{n} \alpha_{ij} \cdot \sigma(\mathbf{w}_j^\intercal \mathbf{x}+b_j)$. 
Also, we define a vector output $g(\mathbf{x}) = \big[ g_1(\mathbf{x}) \, g_2(\mathbf{x}) \cdots \, g_m(\mathbf{x}) \big]$, where each $g_i(\mathbf{x})$ is one of outputs of network before adding a \textit{softmax} layer.
Furthermore, after adding a \textit{softmax} layer to the above outputs, the $i$-th output of the network after the \textit{softmax} layer may be represented as:
$$\mbox{\textup{softmax}}\big(g(\mathbf{x})\big)_i = \frac{\exp\big({g_i(\mathbf{x})}\big)}{\sum_{i=1}^{m}\exp\big({g_i(\mathbf{x})}\big)}
\;\;\;\; (\forall i = 1, 2, \cdots, m)
$$

\section{Main Results}
\label{Results}

If we use the ReLU activation function for all hidden nodes in a neural network in  Figure \ref{single_output_network}. We can prove that it still maintains the universal approximation property  as long as the number of hidden nodes is large enough. 

\begin{theorem} [Universal Approximation with ReLU]\label{theorem-relu-universal}
When ReLU is used as the activation function, the output of a single-hidden-layer neural network
\begin{equation} \label{eq-ReLU-sum-form}
g(\mathbf{x})=\sum_{j=1}^{n} \alpha_j \cdot \mbox{\textup{ReLU}}(\mathbf{w}_j^\intercal \mathbf{x}+b_j)
\end{equation}
in dense in ${L}^1(I_d)$ if $n$ is large enough. In other words, given any function $f(\mathbf{x}) \in{L}^1(I_d)$ and any arbitrarily small positive number $\epsilon>0$, there exists a sum  of the above form $g(\mathbf{x})$, for which:
\begin{equation}
\Vert g-f \Vert_{1} = \int_{I_d}{\Big|g(\mathbf{x}) -f(\mathbf{x}) \Big|}d\mathbf{x} <\epsilon
\end{equation}
\end{theorem}

\textbf{Proof:}

\begin{figure}[h!]
\centering
  \includegraphics[width=0.35\linewidth]{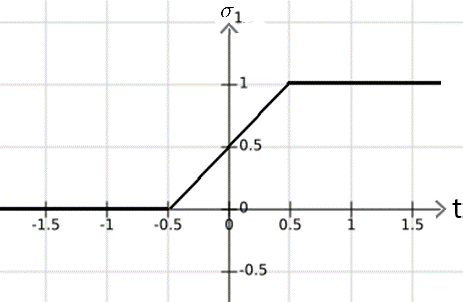}
  \caption{A special activation function $\sigma_1(t)$ as constructed in eq.(\ref{eq-sigma1-function}).}
  \label{sigma}
\end{figure}

First of all, we construct a special activation function use the ReLU function as: 
$$\sigma_1(t) = \mbox{ReLU}(t+0.5)-\mbox{ReLU}(t-0.5)$$
Obviously, $\sigma_1(t)$ is a piece-wise linear function given as follows: 
\begin{equation} \label{eq-sigma1-function}
\sigma_1(t) = \left\{\begin{array}{ll}
        0 & \mbox{ for } t \leq -0.5\\
        t+0.5 & \mbox{ for } -0.5<t<0.5\\
        1 & \mbox{ for } t \geq 0.5
        \end{array}
        \right.
\end{equation}
The above function $\sigma_1(t)$ is also plotted in Figure \ref{sigma}.

Since $\sigma_1(t)$ is a bounded measurable sigmoidal function, based on Theorem 4 in \cite{Cybernko89}, the sum 
$$
\hat{g} (\mathbf{x})=\sum_{j=1}^{n} \alpha_j \cdot \sigma_1(\mathbf{w}_j^\intercal \mathbf{x}+b_j)
$$
is dense in ${L}^1(I_d)$.
In other words, given any function $f(\cdot)\in{L}^1(I_d)$ and $\epsilon>0$, there exists a sum $\hat{g}(\mathbf{x})$ as above, for which:
$$
\int_{I_d} \Big| \hat{g} (\mathbf{x}) - f(\mathbf{x})  \Big| d\mathbf{x} =
\int_{I_d}{\Big|\sum_{j=1}^{n} \alpha_j \sigma_1(\mathbf{w}_j^\intercal \mathbf{x} +b_j) - f(\mathbf{x}) \Big|}d\mathbf{x}<\epsilon
$$
Substituting $\sigma_1(t) = \mbox{ReLU}(t+0.5)- \mbox{ReLU}(t-0.5)$ into the above equation, we have:
$$
\int_{I_d} \Big|\sum_{j=1}^{n} \alpha_j \cdot \mbox{ReLU}(\mathbf{w}_j^\intercal \mathbf{x}+b_j+0.5)
-\sum_{j=1}^{n}\alpha_j \cdot \mbox{ReLU}(\mathbf{w}_j^\intercal \mathbf{x} +b_j-0.5)-f(\mathbf{x})\Big|d\mathbf{x}<\epsilon
$$
We can further re-arrange it into:
\begin{equation} \label{eq-relu-sum}
\int_{I_d}{\Big|\sum_{j=1}^{2n} \tilde{\alpha}_j \mbox{ReLU}(\tilde{\mathbf{w}}_j^\intercal \mathbf{x} +\tilde{b}_j)-f(\mathbf{x}) \Big|}d\mathbf{x}<\epsilon
\end{equation}
where:
\[
\tilde{\alpha}_j = \left\{\begin{array}{ll}
        {\alpha}_j & \mbox{ for } 1\leq j\leq n\\
        -{\alpha}_j & \mbox{ for } n < j\leq 2n
        \end{array}
        \right.
\]

\[
    \tilde{\mathbf{w}}_j = \left\{\begin{array}{ll}
		{\mathbf{w}}_j  & \mbox{ for } 1\leq j\leq n\\
		{\mathbf{w}}_{j-n} & \mbox{ for } n< j\leq 2n
	    \end{array}
        \right.
        \qquad
    \tilde{b}_j= \left\{\begin{array}{ll}
        {b}_j+0.5 & \mbox{ for } 1\leq j\leq n\\
        {b}_j-0.5 & \mbox{ for } n< j\leq 2n
        \end{array}\right.
\]
Given any function $f(\cdot) \in {L}^1(I_d)$ , we have found a sum 
$$
g(\mathbf{x}) = \sum_{j=1}^{2n} \tilde{\alpha}_j \mbox{ReLU}(\tilde{\mathbf{w}}_j^\intercal \mathbf{x} +\tilde{b}_j)
$$
as shown in eq.(\ref{eq-relu-sum}) that satisfies $||g-f||_1<\epsilon$ for any small $\epsilon>0$, therefore the proof is completed.
$\hspace*{\fill} {\hfill\blacksquare}$

In the following, let us extend Theorem \ref{theorem-relu-universal} to a ReLU neural network with multiple outputs without using any \textit{softmax} output layer. 

\begin{corollary} \label{corollary-multi-outputs}
Given any vector-valued function $f(\mathbf{x}) = \big[ f_1(\mathbf{x}) \, f_2(\mathbf{x}) \cdots \, f_m(\mathbf{x}) \big]$, where each component function $f_i(\mathbf{x})\in{L}^1(I_d)$, there exists a single-hidden-layer neural network yielding a vector output 
$g(\mathbf{x}) = \big[  g_1(\mathbf{x}) \, \cdots \,  g_m(\mathbf{x}) \big] $, each of which is  a  sum of the form:
\begin{equation} \label{eq-multi-output-sum}
    g_i(\mathbf{x})=\sum_{j=1}^{n} \alpha_{ij} \cdot \mbox{\textup{ReLU}}(\mathbf{w}_j^\intercal \mathbf{x} +b_j) \;\;\;\; (\forall i = 1, 2, \cdots, m)
\end{equation}
to satisfy  
$$\Vert g - f \Vert_{1}<\epsilon$$ 
for any arbitrarily small positive number $\epsilon>0$, which is 
equivalent to
$$
\sum_{i=1}^{m} \Vert g_i-f_i \Vert_1 = \sum_{i=1}^{m} \int_{I_d} {|g_i(\mathbf{x})-f_i(\mathbf{x})|}d\mathbf{x}<\epsilon
$$
\end{corollary}

\textbf{Proof:}

We use the mathematical induction on $m$ to prove this theorem. 

For $m=1$, the problem reduces to Theorem 1. Next, we assume that the theorem is correct for any $m > 1$, we will prove that it also holds  for $m+1$. Based on our assumption, given $f_1,...,f_m$ and $\frac{\epsilon}{2}>0$, there exist some sums, i.e 
$g_1(\mathbf{x}),...,g_m(\mathbf{x})$, as eq.(\ref{eq-multi-output-sum}) for which:
$$
\sum_{i=1}^{m} \Vert f_i-g_i \Vert_1 = \sum_{i=1}^{m} \int_{I_d} \Big|g_i(\mathbf{x}) -f_i(\mathbf{x}) \Big|d \mathbf{x}<\frac{\epsilon}{2}
$$
$$
\implies 
\sum_{i=1}^{m} \int_{I_d} \Big|\sum_{j=1}^{n} \alpha_{ij} \mbox{ReLU}(\mathbf{w}_j^\intercal \mathbf{x} +b_j)- f_i(\mathbf{x}) \Big| d \mathbf{x}<\frac{\epsilon}{2}
$$
Now we need to approximate $f_{m+1}(\mathbf{x})$. Based on Theorem \ref{theorem-relu-universal}, there exists a sum $\tilde{g}(\mathbf{x})$ of the form in eq.(\ref{eq-ReLU-sum-form}) for which it satisfies for any given $f_{m+1}(\mathbf{x})$ and $\frac{\epsilon}{2}>0$:
$$
\Vert \tilde{g}-f_{m+1} \Vert_{1} = \int_{I_d}{\Big|\tilde{g}(\mathbf{x}) - f_{m+1}(\mathbf{x})\Big|} d \mathbf{x} < \frac{\epsilon}{2}
$$
$$
\implies
\int_{I_d}{\Big|\sum_{j=1}^{n'} \tilde{\alpha}_j \mbox{ReLU}(\tilde{\mathbf{w}}_j^\intercal \mathbf{x} +\tilde{b}_j) - f_{m+1}(\mathbf{x}) \Big|} d\mathbf{x} < \frac{\epsilon}{2}
$$

Now we can construct a new neural network with $m+1$ outputs 
$\big[ g_1(\mathbf{x}) \; g_2(\mathbf{x}) \cdots  g_m(\mathbf{x}) \; \tilde{g}(\mathbf{x}) \big]$,  and all weights are defined as follows:
\[
\bar{\alpha}_{ij} = \left\{\begin{array}{ll}
        {\alpha}_{ij} & \mbox{ for } 1\leq j\leq n \mbox{, } 1\leq i\leq m\\
        0 & \mbox{ for } n< j\leq n+n' \mbox{, } 1\leq i\leq m\\
        0 & \mbox{ for } 1\leq j\leq n+n' \mbox{, } i=m+1\\
        \tilde{\alpha}_{j-n} & \mbox{ for } n< j\leq n+n' \mbox{, } i=m+1
        \end{array}
        \right.
\]
\[
\bar{\mathbf{w}}_j = \left\{\begin{array}{ll}
		{\mathbf{w}}_j  & \mbox{ for } 1\leq j\leq n\\
		\tilde{\mathbf{w}}_{j-n} & \mbox{ for } n< j\leq n+n'
	    \end{array}
        \right.
        \qquad
        \bar{b}_j= \left\{\begin{array}{ll}
        {b}_{j} & \mbox{ for } 1\leq j\leq n\\
        \tilde{b}_{j-n} & \mbox{ for } n< j\leq n+n'
        \end{array}\right.
\]
Now we can show:
$$
\sum_{i=1}^{m+1} \int_{I_d} {|\sum_{j=1}^{n+n'} \bar{\alpha}_{ij} \mbox{ReLU}(\bar{\mathbf{w}}_j^\intercal \mathbf{x}+\bar{b}_j)-f_i(\mathbf{x})|}d \mathbf{x}
$$
$$
= \sum_{i=1}^{m} \int_{I_d} {\Big|\sum_{j=1}^{n+n'} \bar{\alpha}_{ij} \mbox{ReLU}(\bar{\mathbf{w}}_j^\intercal \mathbf{x}+\bar{b}_j)-f_i(\mathbf{x}) \Big|}d\mathbf{x}
$$
$$
+ \int_{I_d} {\Big|\sum_{j=1}^{n+n'} \bar{\alpha}_{m+1,j} \mbox{ReLU}(\bar{\mathbf{w}}_j^\intercal \mathbf{x}+\bar{b}_j)-f_{m+1}(\mathbf{x}) \Big|}d\mathbf{x}
$$
$$
= \sum_{i=1}^{m} \int_{I_d} {\Big|\sum_{j=1}^{n} {\alpha}_{ij} \mbox{ReLU}({\mathbf{x}}_j^\intercal \mathbf{x} +{b}_j)-f_i(\mathbf{x}) \Big|}d\mathbf{x}
$$
$$
+ \int_{I_d} {\Big|\sum_{j=n+1}^{n+n'} \tilde{\alpha}_{m+1,j-n} \mbox{ReLU}(\tilde{\mathbf{w}}_{j-n}^\intercal \mathbf{x}+\tilde{b}_{j-n})-f_{m+1}(\mathbf{x})\Big|}d \mathbf{x}
$$
$$
< \frac{\epsilon}{2}+\frac{\epsilon}{2}=\epsilon
$$
Therefore, the proof is completed.
$\hspace*{\fill} {\hfill\blacksquare}$

In the following, we will investigate how the \textit{softmax} output layer may affect the approximation capabilities of neural networks. 

\begin{lemma} \label{lemma-softmax}
Given any vector-valued function $f(\mathbf{x}) = \big[ f_1(\mathbf{x}) \, \cdots f_m(\mathbf{x}) \big]$ where each component function  $f_i(\mathbf{x})\in {L}^1(I_d)$, there exists a vector-valued function $g(\mathbf{x}) = \big[ g_1(\mathbf{x}) \, \cdots g_m(\mathbf{x}) \big]$, each $g_i(\mathbf{x})$ is a sum of eq.(\ref{eq-multi-output-sum}), so that their  \textit{softmax} outputs satisfy:
\begin{equation}
\Vert \mbox{\textup{softmax}}(g(\mathbf{x}))_i-\mbox{\textup{softmax}}(f(\mathbf{x}))_i\Vert_1 < \frac{\epsilon}{2} \;\;\;\; (\forall i=1,2,\cdots,m)
\end{equation}
for any small positive value $\frac{\epsilon}{2}>0$,  where $$\mbox{\textup{softmax}}(g(\mathbf{x}))_i = \frac{\exp\big({g_i(\mathbf{x})}\big)}{\sum_{i=1}^{m}\exp\big({g_i(\mathbf{x})}\big)}
\;\;\;\; \mbox{\textup{and}} \;\;\;\;
\mbox{\textup{softmax}}(f(\mathbf{x}))_i = \frac{\exp\big({f_i(\mathbf{x})}\big)}{\sum_{i=1}^{m}\exp\big({f_i(\mathbf{x})}\big)}$$
\end{lemma}

\textbf{Proof:}

First, the \textit{softmax} function is a continuous function everywhere in its domain.\footnote{According to continuity of exp function and positivity of the denominator of $H_i$, the proof of this claim is straightforward and is omitted.} Therefore,  given any $\frac{\epsilon}{2}>0$, there exists a $\delta>0$ such that 
if $\Vert g - f \Vert_{1}<\delta$ then $\Vert \mbox{\textup{softmax}}(g(\mathbf{x})) - \mbox{\textup{softmax}}(f(\mathbf{x})) \Vert _1<\frac{\epsilon}{2}$.

Second, according to  Corollary \ref{corollary-multi-outputs}, for any $\delta>0$ and $f \in \big({L}^1(I_d) \big)^m$, there exists a function $g(\cdot) \in \big({L}^1(I_d) \big)^m$ such that:
$$
\Vert g - f \Vert_{1}<\delta
$$
Putting these two results together completes the proof.
$\hspace*{\fill} {\hfill\blacksquare}$

\begin{definition}[Indicator Function]
We define a vector-valued function $f(\mathbf{x}) = \big[ f_1(\mathbf{x}) \, \cdots \, f_m(\mathbf{x}) \big]$ as an indicator function if it simultaneously satisfies the following two conditions:
\begin{enumerate}
\item $\forall \mathbf{x}$ and $1 \leq i \leq m$:
$
f_i( \mathbf{x}) =
\left\{
\begin{array}{ll}
		1  \\
		0 
\end{array}
\right.
$
\item $\forall \mathbf{x} : \sum_{i=1}^{m} f_i(\mathbf{x})=1$
\end{enumerate}
\end{definition}

In other words, for any input $\mathbf{x}$, an \textit{indicator function} will yield one but only one '1' in a component and '0' for all remaining components. Obviously, an indicator can represent mutually-exclusive class labels in a multiple-class pattern classification problem. On the other hand, the class labels from any such pattern classification problem may be represented by an \textit{indicator function}.

\begin{theorem} [Approximation Capability of Softmax] \label{theorem-approx-softmax}
Given any indicator function $f(\mathbf{x})= \big[ f_1(\mathbf{x}) \, \cdots \, f_m(\mathbf{x}) \big]$, assume each component function $f_i(\mathbf{x})\in {L}^1(I_d)$ ($\forall i =1,\cdots,m$), 
there exist some sums as eq.(\ref{eq-multi-output-sum}), i.e. $g(\mathbf{x})= \big[ g_1(\mathbf{x}) \, \cdots \, g_m(\mathbf{x}) \big]$,
such that their corresponding \textit{softmax} outputs satisfy: 
\begin{equation}
\Vert \mbox{\textup{softmax}}\big(g(\mathbf{x})\big)_i -f_i(\mathbf{x})\Vert_1 < \epsilon  \;\;\;\; (\forall i = 1, 2, \cdots m)
\end{equation}
for any small $\epsilon >0$, 
where $\mbox{\textup{softmax}}\big(g(\mathbf{x})\big)_i = \frac{\exp\big({g_i(\mathbf{x})}\big)}{\sum_{i=1}^{m}\exp\big({g_i(\mathbf{x})}\big)}$.
\end{theorem}

\textbf{Proof:}

First, we define a new vector-valued function $f'(\mathbf{x}) = \big[f'_1(\mathbf{x}) \,  f'_2(\mathbf{x})  \cdots f'_m(\mathbf{x}) \big]$, where each component function is defined as functions $f'_i(\mathbf{x})=\frac{2m}{\epsilon}(f_i(\mathbf{x})-0.5)$ for all $1\leq i \leq m$. Note that $f'_i(\mathbf{x}) \in {L}^1(I_d)$ due to the assumption $ f_i(\mathbf{x}) \in {L}^1(I_d)$. According to the triangular inequality, for any $1\leq i \leq m$, we have:
$$
\Vert \mbox{\textup{softmax}}\big(g(\mathbf{x})\big)_i -f_i(\mathbf{x})\Vert _{1} \leq 
\Vert \mbox{\textup{softmax}}\big(g(\mathbf{x})\big)_i - \mbox{\textup{softmax}}\big(f'(\mathbf{x})\big)_i \Vert_{1} + 
\Vert \mbox{\textup{softmax}}\big(f'(\mathbf{x})\big)_i - f_i(\mathbf{x}) \Vert_{1}
$$

Second, based on Lemma \ref{lemma-softmax}, for our any given $f'_i(\mathbf{x})$, there exists a function  $g(\mathbf{x}) = \big[ g_1(\mathbf{x}) \, \cdots  \,g_m(\mathbf{x}) \big]$, each of which is in the form of eq.(\ref{eq-multi-output-sum}), to ensure 
$\Vert \mbox{\textup{softmax}}\big(g(\mathbf{x})\big)_i - \mbox{\textup{softmax}}\big(f'(\mathbf{x})\big)_i\Vert_{1}<\frac{\epsilon}{2}$. 

Next, we just need to show that for any indicator function $f(\mathbf{x} ) = \big[f_1(\mathbf{x} ) 
\, \cdots \, f_m(\mathbf{x} ) \big]$, where each $f_i(\mathbf{x} )\in{L}^1(I_d)$, we have $\Vert \mbox{\textup{softmax}}\big(f'(\mathbf{x})\big)_i - f_i(\mathbf{x}) \Vert_{1}\leq\frac{\epsilon}{2}$ holds for any $i$. In order to prove this, we first have:
\begin{eqnarray}
& & \Vert \mbox{\textup{softmax}}\big(f'(\mathbf{x})\big)_i - f_i(\mathbf{x}) \Vert_{1} =   \int_{I_d}^{}\Big|\mbox{\textup{softmax}}\big(f'(\mathbf{x})\big)_i - f_i(\mathbf{x}) \Big|d \mathbf{x}
\nonumber \\
& = & \int_{I_d,f_i=1}\Big|\mbox{\textup{softmax}}\big(f'(\mathbf{x})\big)_i - f_i(\mathbf{x}) \Big|d \mathbf{x} + \int_{I_d,f_i=0}^{}\Big|\mbox{\textup{softmax}}\big(f'(\mathbf{x})\big)_i - f_i(\mathbf{x}) \Big|d \mathbf{x}
\nonumber \\
& = & \int_{I_d,f_i=1}\Big|\frac{\exp\big(\frac{2m}{\epsilon}(f_i(\mathbf{x})-0.5))} {\sum_{i=1}^{m}\exp\big(\frac{2m}{\epsilon}(f_i(\mathbf{x})-0.5)\big)}-1 \Big|d\mathbf{x}
+ \int_{I_d,f_i=0} \Big|\frac{\exp\big(\frac{2m}{\epsilon}(f_i(\mathbf{x})-0.5) \big)} {\sum_{i=1}^{m}\exp\big(\frac{2m}{\epsilon}(f_i(\mathbf{x})-0.5)\big)}| d\mathbf{x} \nonumber
\end{eqnarray}
Based on  the properties of the indicator function $f(\mathbf{x} )$, we further derive:
\begin{eqnarray}
& = & \int_{I_d,f_i=1}\Big|\frac{\exp(\frac{\epsilon}{m})- \exp(\frac{\epsilon}{m}) - (m-1)\exp(-\frac{\epsilon}{m})} {\exp(\frac{\epsilon}{m}) + (m-1)\exp(-\frac{\epsilon}{m})} \Big|
d\mathbf{x} \nonumber \\
& + & \int_{I_d,f_i=0}\Big|\frac{\exp(-\frac{\epsilon}{m})} {\exp(\frac{\epsilon}{m}) + (m-1)\exp(-\frac{\epsilon}{m})} \Big|d \mathbf{x}
\nonumber \\
& < & \int_{I_d,f_i=1}\Big|\frac{(m-1)\exp(-\frac{\epsilon}{m})} {\exp(\frac{\epsilon}{m})} \Big|d\mathbf{x}  + \int_{I_d,f_i=0} \Big|\frac{\exp(-\frac{\epsilon}{m})} {\exp(\frac{\epsilon}{m})} \Big| d \mathbf{x}
\nonumber \\
& \leq & (m-1)\exp\big(-\frac{2\epsilon}{m})+\exp(-\frac{2\epsilon}{m} \big) = m\exp \big(-\frac{2\epsilon}{m} \big) \leq m\frac{\epsilon}{2m} = \frac{\epsilon}{2} \nonumber
\end{eqnarray}
where the last step at above follows the inequality  $\exp(-x)\leq \frac{1}{x}$ for $x>0$. 

Finally, for any $1 \leq i \leq m$,  we have
$$
\Vert \mbox{\textup{softmax}}\big(g(\mathbf{x})\big)_i -f_i(\mathbf{x})\Vert _{1} \leq
\frac{\epsilon}{2} + \frac{\epsilon}{2} = \epsilon
$$
Therefore, the proof is completed. 
$\hspace*{\fill} {\hfill\blacksquare}$

Obviously, Theorem \ref{theorem-approx-softmax} indicates that a sufficiently-large single-hidden-layer feed-forward neural network with a softmax layer can approximate well any indicator function if the target function belongs to ${L}^1(I_d)$.

\section{Conclusions}
\label{Conclusions}

In this work, we have studied the approximation capabilities of the popular ReLU activation function and softmax output layers in neural networks for pattern classification. We have first shown in Theorem \ref{theorem-relu-universal} that a large enough neural network using the ReLU activation function is a universal approximator in $L^1$. Furthermore, we have extended the result to multi-output neural networks and have proved that it can approximate any target function in $L^1(I_d)$. Next, we have proved in Theorem \ref{theorem-approx-softmax} that a sufficiently large neural network can approximate well any indicator target function in $L^1$, which is equivalent to mutually-exclusive class labels in any realistic multiple-class pattern classification tasks. At last, we also want to note that the result in Theorem \ref{theorem-approx-softmax} is also applicable to many other activation functions in \cite{Cybernko89,Kurt1991251} since we do not use the property of the ReLU activation function in the proof. 

\medskip

\bibliography{ReLU_Softmax_theory}
\bibliographystyle{unsrt}

\end{document}